\documentclass{article}

\PassOptionsToPackage{numbers, compress}{natbib}
\usepackage[preprint]{neurips_2019}

\usepackage[utf8]{inputenc} % allow utf-8 input
\usepackage[T1]{fontenc}    % use 8-bit T1 fonts
\usepackage{hyperref}       % hyperlinks
\usepackage{url}            % simple URL typesetting
\usepackage{booktabs}       % professional-quality tables
\usepackage{amsfonts}       % blackboard math symbols
\usepackage{nicefrac}       % compact symbols for 1/2, etc.
\usepackage{microtype}      % microtypography

\usepackage{graphicx}
\usepackage{amsmath}
\usepackage{subcaption}
\usepackage{appendix}

\title{Few-Shot Abstract Visual Reasoning With Spectral Features}

\author{
  Tanner Bohn \\%\thanks{Use footnote for providing further information
  Department of Computer Science\\
  Western University\\
  London, ON, Canada \\
  \texttt{tbohn@uwo.ca} \\
   \And
   Yining Hu \\
   Department of Computer Science\\
  Western University\\
  London, ON, Canada \\
  \texttt{yhu534@uwo.ca} \\
   \AND
   Charles X. Ling \\
   Department of Computer Science\\
   Western University\\
   London, ON, Canada \\
   \texttt{charles.ling@uwo.ca} \\
}

\begin{document}

\maketitle

\begin{abstract}
We present an image preprocessing technique capable of improving the performance of few-shot classifiers on abstract visual reasoning tasks. Many visual reasoning tasks with abstract features are easy for humans to learn with few examples but very difficult for computer vision approaches with the same number of samples, despite the ability for deep learning models to learn abstract features. Same-different (SD) problems represent a type of visual reasoning task requiring knowledge of pattern repetition within individual images, and modern computer vision approaches have largely faltered on these classification problems, even when provided with vast amounts of training data. We propose a simple method for solving these problems based on the insight that removing peaks from the amplitude spectrum of an image is capable of emphasizing the unique parts of the image. When combined with several classifiers, our method performs well on the SD SVRT tasks with few-shot learning, improving upon the best comparable results on all tasks, with average absolute accuracy increases nearly 40\% for some classifiers. In particular, we find that combining Relational Networks with this image preprocessing approach improves their performance from chance-level to over 90\% accuracy on several SD tasks.
%The success of this approach opens the door to solving a wider variety of real-word computer vision problems requiring abstract reasoning.

% TODO: utilize structure from this ICCV-19 SUMs abstract
%We present a conceptually simple image preprocessing step for few-shot learning of traditionally difficult same-different (SD) visual reasoning tasks. 

%SD problems requires identifying repeated patterns in an image, and are often easy for humans to solve with few examples while deep learning approaches have required millions to obtain above chance-level performance. 

%In this paper we show that a particular type of visual saliency map, which we term a self-uniqueness map, provides useful information for downstream classifiers in learning to solve SD problems. 

%Utilizing self-uniqueness maps, we improve upon the best comparable few-shot learning results on all of the SVRT SD tasks with an average absolute accuracy increase above 10\%. 

%We find that while providing Relational Networks with raw SD images achieves chance-level accuracy, with SUMs, the performance on SD tasks increases by as much as 30\% on select problems. 

%This work suggests that self-uniqueness maps are an intuitive and promising feature to use in solving related abstract reasoning tasks.
\end{abstract}

\section{Introduction}
\label{sec:introduction}

%ICML 2019 formatting instructions: %https://media.neurips.cc/Conferences/ICML2019/Styles/example_paper.pdf 

%http://approximatelycorrect.com/2018/01/29/heuristics-technical-scientific-writing-machine-learning-perspective/

The field of artificial intelligence has slowly enveloped human skills ranging from those requiring formal reasoning to those requiring flexibility and intuition (such as image recognition or natural language understanding). The recognition of highly abstract concepts given few examples is a ubiquitous human skill that has yet to see significant progress. In particular, flexible machine learning techniques have not yet been found which can solve a certain type of visual reasoning task known as same-different (SD) problems given only few examples~\citep{stabinger201625, ricci2018not}. Solving these problems requires reasoning about the similarity between patterns located within the same image, something humans perform with ease. In this work we present a conceptually simple image transformation which can be combined with few-shot image classifiers to perform well at these tasks.

\begin{figure}[ht]
\centering
\includegraphics[width=\columnwidth]{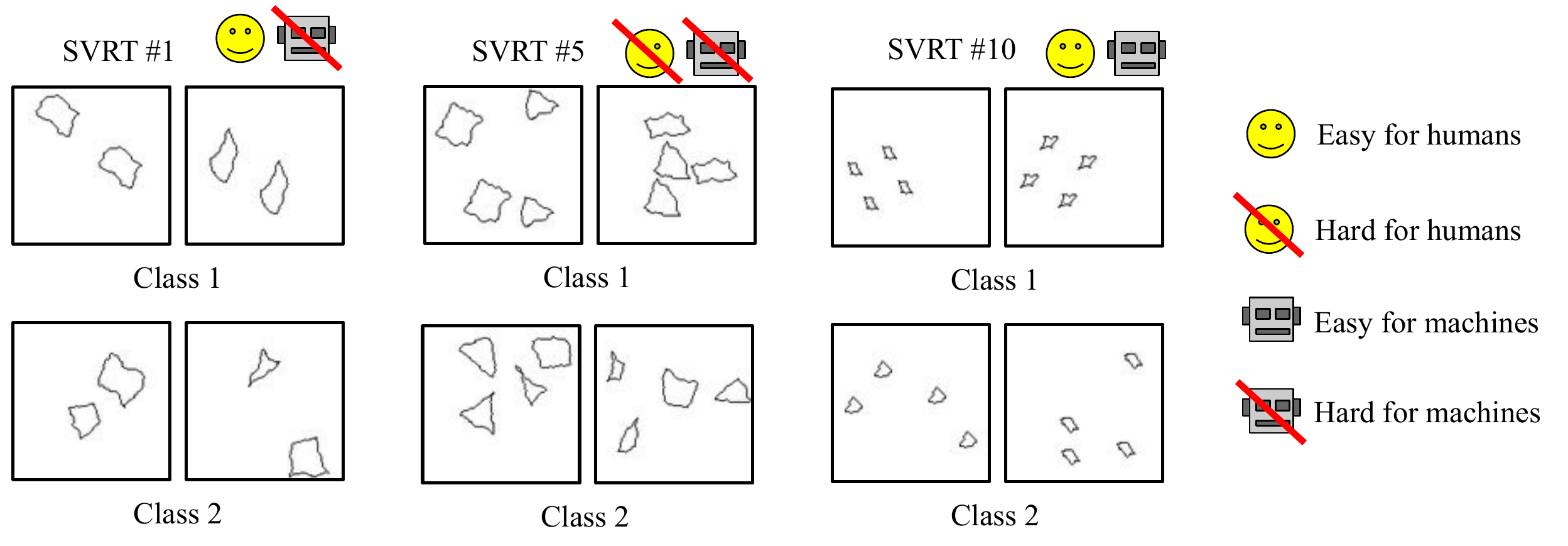}
\caption{Representative examples from the SVRT task set. For SVRT \#1, class 1 contains two identical figures, while images from class 2 contain two unique figures. For SVRT \#5, class 1 contains two pairs of identical figures, while class 2 contains four unique figures. For SVRT \#22, class 1 contains three identical aligned shapes, while in class 2, the shapes are not identical. For all three of these tasks, deep learning approaches have been unable to achieve good performance, even with large amounts of training data. In contrast, our approach is capable of achieving high accuracies with only 10 labelled samples.}
\label{fig:svrt_examples}
\end{figure}

There are many potential sources of abstract visual reasoning tasks one could study. These include Bongard problems (given 6 labelled images, try find the rule separating them)~\citep{bongard1967problem}, Raven's Progressive Matrices (given observed sequences of tiles, identify the missing tile)~\citep{raven1938raven}, the CLEVR dataset (answer questions based on a scene of multiple objects)~\citep{johnson2017clevr}, and SVRT problems (given N labelled images, classify unseen images)~\citep{fleuret2011comparing}.  Although the SVRT lacks the variation of hand-drawn Bongard Problems (23 tasks vs. 200), it contains a variety of highly abstract tasks while being procedurally generated. Unlike Raven's Progressive Matrices or CLEVR tasks which have seen significant progress~\citep{santoro2018measuring, hudson2018compositional}, SVRT tasks have seen only partial success. It is worth noting that while creating a program to automatically solve these types of problems may often be easy (the images in these tasks are simple, consisting of simple objects on a solid white background), the difficulty comes from finding an approach capable of learning to solve the problems with a minimal amount of expert knowledge or feature engineering.

The 23 SVRT tasks can be split into two groups, based on the type of patterns separating the two classes: spatial relation (SR) problems (ex. shapes in a line vs. not in a line) and SD problems (ex. two pairs of unique shapes vs. two pairs of identical shapes)~\citep{stabinger201625}. Previous attempts at these tasks have shown that convolutional neural networks, a staple of image classification research, are capable of solving SR problems given enough training data (20K in~\citep{stabinger201625} and 1 million in~\citep{ricci2018not}), but the SD problems have proven to be more difficult. Perhaps unexpectedly, even relational networks~\citep{santoro2017simple}, which perform well on other reasoning tasks such as CLEVR, have demonstrated great difficulty in learning to solve same-different problems, requiring millions of training samples to achieve above chance-level performance~\citep{ricci2018not}.

\begin{figure}[ht]
\centering
\includegraphics[width=0.7\columnwidth]{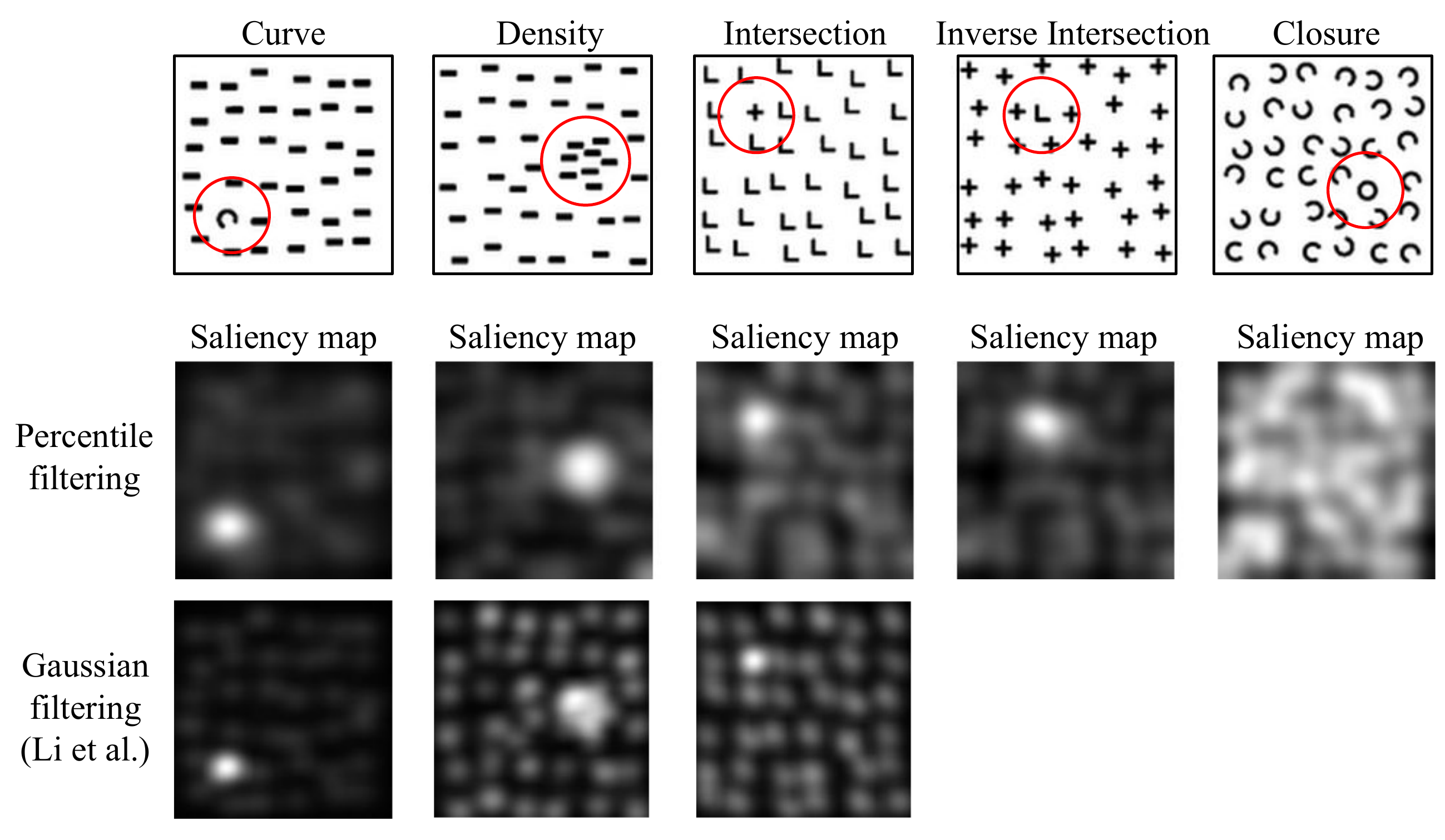}
\caption{A comparison of saliency maps produced by percentile filtering and Gaussian filtering~\citep{li2013visual} on psychological test patterns from~\citep{hou2007saliency}. The expected salient locations are circled. As demonstrated by~\citep{hou2007saliency, seo2009static, guo2008spatio}, the closure test proves to be difficult for many automated methods (this test was not reported by~\citep{li2013visual}). Following~\citep{hou2007saliency, li2013visual}, we square the raw saliency map to increase intensity contrast and apply a Gaussian filter to smooth the highlighted regions.}
\label{fig:gaze_tests}
\end{figure}

Our new approach is inspired by an interesting point of overlap between image processing and visual saliency, namely, Fourier transforms. In image processing, it is well-known that unwanted periodic noise in images can be removed by manually zeroing-out corresponding peaks in the amplitude spectrum of the Fourier transform with so-called 'notch' filters. Instead of manually removing peaks, a functionally similar approach is to smooth peaks, which can be done automatically. In~\citep{li2013visual}, the authors point out that a Gaussian filtering operation on amplitude spectra is an elegant way to compute visual saliency maps. This technique works because non-salient parts of an image are often those which are frequently occurring, leading to peaks in the amplitude spectrum. The effectiveness of this simple approach for producing saliency maps can be seen in the bottom row of Figure \ref{fig:gaze_tests}. In this work, we apply this insight to solve SD tasks. Duplicated figures in SVRT images that undergo this amplitude spectrum filtering process are partially removed, whereas unique figures are largely untouched. This provides easily usable information for convolutional neural-network-based approaches to learn to classify images from SD tasks.

We observe that using these filtered spectra in place of the raw images, we are able to improve the state-of-the-art at few-shot learning SVRT SD tasks by an average of nearly 30\%. By working towards solving these abstract visual problems in a simple and interpretable way, we demonstrate how analogous problems in other areas of machine learning may be approached.

% https://arxiv.org/pdf/1605.01999.pdf
%~\citep{li2013visual}
%   - spikes in amplitude correspond to repeated patterns (pg 4)
%   - Suppressing Repeated Patterns Using Spectral Filtering
%   - (note that this work also skips the closure test)

The main contributions of this work are thus:
\begin{itemize}
    \item We describe a novel approach for solving same-different visual reasoning tasks which exploits insights from visual saliency (Section~\ref{sec:our_model}).
    \item We establish the performance of several popular algorithms for few-shot learning on SVRT tasks (Section~\ref{sec:results}).
    \item We experimentally demonstrate that our approach allows for achieving state-of-the-art performance on all of the SVRT same-different tasks with few-shot learning (Section~\ref{sec:obs_sd}).
\end{itemize}

%The outline for the rest of the paper is as follows. In Section \ref{sec:related} we provide an overview the related work, including research on same-different tasks and visual saliency maps. In Section \ref{sec:our_model} we describe our new approach for same-different problems. In Section \ref{sec:setup} we discuss the experimental setup, including the dataset used and how hyperparameters of our models are tuned. In Section \ref{sec:results} we provide experimental results comparing our new approach to published results as well as several baseline feature types. Section \ref{sec:conclusion} concludes the paper.

%Similar to last into -- just modify part about new method -- mention its application to removing patterned noise (include figure to verify that it works). We highlight a subset of the problems that are particularly difficult… show that they’re hard even with more data (this is analogous to what previous work has shown… slowly cutting away at the problems that are truly difficult)

\section{Related Work}
\label{sec:related}

In this section we first focus on previous work done on SD problems in particular, then provide an overview of relevant work on visual saliency maps.

\subsection{Same-Different Visual Reasoning Tasks} 

%Perhaps due to the difficulty in learning to solve SD tasks, most work in this area has focused on examining how and why machine learning approaches falter on these tasks. 
With the aim of highlighting the difference in performance between humans and machines at solving visual reasoning tasks, Fran{\c{c}}ois Fleuret et al. introduced the SVRT tasks in 2011~\citep{fleuret2011comparing}. With their set of 23 binary classification tasks, the authors demonstrated that humans are much more proficient than a set of standard machine learning image classification techniques chosen at the time. In their few-shot experiments (only 10 training samples), the performance on most of the 23 tasks hovered around random. With 10,000 training images however, their best models were able to obtain 81\% accuracy on SD tasks and 88\% accuracy on SR tasks.

Studying the SVRT tasks with more modern computer vision approaches, Sebastian Stabinger et al.~\citep{stabinger201625} trained LeNet~\citep{lecun1989backpropagation} and GoogLeNet~\citep{szegedy2015going} CNNs on the tasks. They found that near-perfect performance was achievable on roughly half of the tasks, with the other half being significantly more difficult (near-random). By observing the abstract concepts required to solve each SVRT task, they noticed that this easy-difficult split closely corresponded to whether or not tasks required same-different comparisons, with a couple exceptions (they determined that a couple SD problems could be solved by exploiting simple pixel distribution patterns). The authors find that both CNN models perform very similarly, achieving perfect or near-perfect accuracies on the spatial-relation problems but poor performance on remaining tasks, despite training each network on 20,000 images. In~\citep{ricci2018not}, the authors use a minimal synthetic SVRT-like task for deeper exploration. They show that relational networks (RNs)~\citep{santoro2017simple} have trouble learning to solve these problems, requiring several million training samples before achieving above chance-level performance. %An inverse graphics route to solving visual reasoning tasks was taken by Kevin Ellis et al.~\citep{ellis2015unsupervised}. However, despite demonstrating strong performance, program synthesis for some images taking hundreds of seconds.

%%%%%%%%%%%%%%%%%%%%%%%%%%%%%%%%%%%%%%%%%%%%%%%%%%%%%%%%%%%%%
\subsection{Saliency Maps} 
\label{sec:related_sal}

The aim of a visual saliency map is to indicate where a human is likely to gaze when looking at an scene, and these maps have proven to be useful for a wide variety of applications. Knowing where humans will attend in an image allows for such interesting applications as content aware image resizing~\citep{suh2003automatic, avidan2007seam}, segmentation of salient objects~\citep{rahtu2010segmenting}, aiding general objection-recognition~\citep{rutishauser2004bottom}, and video summarization~\citep{ma2005generic}. 

Methods for calculating visual saliency can roughly be grouped by their underlying assumptions on the meaning of saliency. One common idea is that saliency is "an anomaly with respect to context"~\citep{wang2011image}. Separate approaches thus exist for various interpretations of "context". Approaches that use the local context around a pixel to determine saliency are often based on low-level visual features like colour, colour intensity, edge orientation, or texture~\citep{goferman2012context}. Another set of approaches use a global context, where the saliency at a location depends on the entire image. One technique is to explicitly compare every patch in the image to a representative set of other patches. Those patches that are unique will have low cosine-similarities to other patches~\citep{seo2009static, goferman2012context}. 

Another group of approaches most relevant to the present paper has made use of operations on the Fourier transform of images. Adopting notation from~\citep{li2013visual}, we consider $ f(x, y) \xrightarrow{\mathcal{F}} \mathcal{F}(f)(u, v)$ to be the mapping of image $f(x,y)$ to the frequency domain, where $\mathcal{A}(u, v) = |\mathcal{F}(f)|$ is the amplitude spectrum and $\mathcal{P}(u, v) = angle(\mathcal{F}(f))$ is the phase spectrum. Xiaodi Hou and Liqing Zhang~\citep{hou2007saliency} proposed the idea that novelty in images is represented in the amplitude spectra of images as 'residuals': 
\begin{equation}
\mathcal{R}(u, v) = \mathcal{L}(u, v) - h_{n} \star \mathcal{L}(u, v),
\end{equation}
where $\mathcal{L}(u,v) = \log{\mathcal{A}(u, v)}$ and $h_{n}$ is an $n \times n$ low-pass filter convolved over $\mathcal{L}(u, v)$. To produce the saliency map, $\mathcal{S}(x, y)$, the residual is then used in place of the original amplitude:
\begin{equation}
    \mathcal{S}(x, y) = \mathcal{F}^{-1}[\mathcal{R}(u, v) \cdot exp(i \cdot \mathcal{P}(u, v)) ].
\end{equation}
To achieve more visually pleasing results, they additionally square the elements of $\mathcal{S}(x,y)$ and smooth it with a low-pass filter. Soon after this work,~\citep{guo2008spatio} showed that similar results could be achieved by simply reconstructing the image using the phase spectrum:
\begin{equation}
\label{eq:only_phase_1}
    \mathcal{S}(x, y) = \mathcal{F}^{-1}[1(u, v) \cdot exp(i \cdot \mathcal{P}(u, v)) ].
\end{equation}
In~\citep{li2013visual}, the authors suggest that these two approaches achieve very similar visual results because the residual of the amplitude spectrum is very similar to a plane (i.e. $1(u,v)$) and reconstructing images with only phase information functions similar to a gradient enhancement, highlighting object boundaries and textured parts of the image. This property leaves the methods ill-suited for identifying large salient regions or salient regions in front of noisy backgrounds.

Guided by the fact that repeated patterns (also called "non-salient" patterns) correspond to spikes in the amplitude spectrum, the authors of~\citep{li2013visual} reason that suppressing these peaks corresponds to leaving only the salient parts of an image. To perform this suppression so that sharper spikes are reduced more, a low-pass Gaussian filter, $g$, is used to obtain the smoothed amplitude spectrum:
\begin{equation}
\label{eq:only_phase_2}
    \mathcal{A}_{\mathcal{S}}(u, v) = \mathcal{A}(u, v) \star g.
\end{equation}
The saliency map can then be calculated with:
\begin{equation}
\label{eq:sal_map_smooth}
    \mathcal{S}(x, y) = \mathcal{F}^{-1}[\mathcal{A}_{\mathcal{S}}(u, v) \cdot exp(i \cdot \mathcal{P}(u, v)) ].
\end{equation}
This method is the most similar to our work described in the next section, One main difference is that we additionally consider a percentile filer. Additionally, instead of applying the inverse Fourier transform, we experimentally observe (Section~\ref{sec:results_hyperparams}) that training image classifiers directly on the filtered amplitude spectra leads to improved results on SD problems.

More recent CNN based approaches to saliency also exist~\citep{kummerer2014deep,kruthiventi2017deepfix}, however, work by \citep{stabinger201625, ricci2018not} suggests they do not transfer well to the abstract SVRT tasks despite the excellent performance of these approaches on calculating saliency maps.

\section{Our Approach}
\label{sec:our_model}

\begin{figure}[htp]
\centering
\includegraphics[width=0.8\columnwidth]{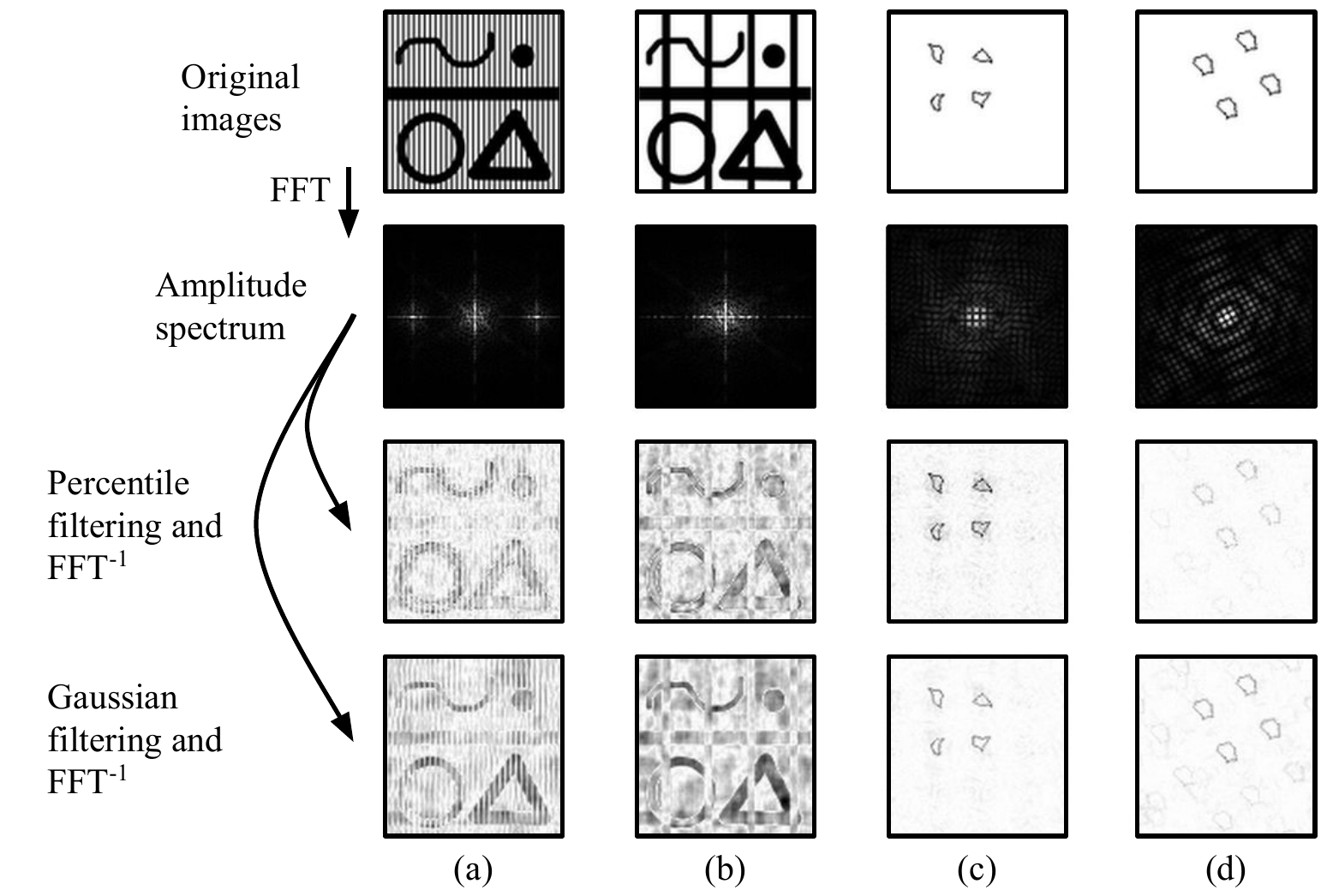}
\caption{A demonstration of how non-unique parts of an image are affected by amplitude spectrum filtering. In the top row, columns (a) and (b) are images with varying scales of a repeated vertical lines and columns (c) and (c) are positive and negative samples from SVRT \#15. The second row contains the amplitude spectra. The third and fourth rows contain the figures reconstructed with the percentile- and Gaussian-filtered amplitude spectra respectively. Note how the duplicated shapes are partially removed following filtering -- for columns (c) and (d) containing SVRT samples, this effect is more clear with the percentile filter. The images in row 2 have undergone contrast enhancement for better printing quality. Rows 3 and 4 have been individually normalized so that their intensities lie in the range $[0, 1]$.}
\label{fig:peak_removal}
\end{figure}

The core of our SD problem-solving approach is based on the insight that peaks in the amplitude spectrum of an image correspond to the non-unique parts of that image, and removing these peaks corresponds to removing the non-unique parts of the image~\citep{li2013visual} (demonstrated in Figure~\ref{fig:peak_removal}). While deep convolutional neural networks have been able to solve a wide variety of visual tasks given only the raw images, they have thus far been unable to solve SD tasks this way~\citep{stabinger201625, ricci2018not}. A motivation of our model is thus to find a simple transformation of the problem images such that when combined with CNN-based approaches, the classifiers are capable of extracting the necessary information to solve SD problems. Consider SVRT \#1 from Figure~\ref{fig:svrt_examples} for example. If an image transformation were capable of making the non-unique figure outlines in class 1 lighter than the unique figure outlines in class 2, then a CNN classifier trained with gradient-descent would have no trouble identifying the relevant visual feature (namely, intensity). This is precisely what amplitude spectrum filtering allows us to do. The main way our primary approach differs from this is that instead of training the classifier on the inverse Fourier transform using the filtered amplitude, we provide the filtered amplitude spectrum directly to the classifier. In Section~\ref{sec:results_hyperparams} we examine how using the filtered amplitude differs from using the inverse Fourier transform with the filtered amplitude.

In the remainder of this section, we first discuss the intuition of how amplitude filtering works to detect uniqueness and the difference between filtering methods we consider. Second, we discuss how these amplitude spectra fit into the rest of our problem-solving pipeline.

% https://staff.fnwi.uva.nl/r.vandenboomgaard/IPCV20172018/LectureNotes/IP/LocalOperators/percentilefilter.html

\subsection{Amplitude Spectra Filtering}

%we first discuss the intuition of how amplitude filtering works to detect uniqueness and the difference between filtering methods we consider

	\textbf{Removing non-uniqueness.} Demonstrations of how amplitude filtering affects non-uniqueness in an image are provided in Figure~\ref{fig:peak_removal}. In columns (a) and (b), we can see an image with several unique figures and different scales of repeated (i.e. non-unique) vertical bars superimposed. In the amplitude spectra, these vertical bars correspond to the sharp peaks located symmetrically about the y-axis observable in the second row of the figure. By applying a filter to smooth or remove these peaks before reconstructing the image, we remove the cause of those peaks. As demonstrated, this works even when there are relatively few instances of the repeated shape (column (b)).

	\textbf{Filtering methods.} To perform the amplitude spectrum filtering for our model, we consider Gaussian filtering (which was used in~\citep{li2013visual}) and (primarily) percentile filtering. Gaussian filtering of a matrix is performed by convolving a kernel whose values approximate a 2D Gaussian function:
	\begin{equation}
		G(x, y) = \frac{1}{2 \pi \sigma^{2}} e^{-\frac{x^{2} + y^{2}}{2 \sigma ^{2}}},
	\end{equation}
	where $x$ and $y$ are respectively the horizontal and vertical distances from the origin and $\sigma$ is the standard deviation. For each position, $\mathbf{u}$, in the amplitude spectrum of a discrete Fourier transform, the $p$-percentile filtered value is:
	\begin{equation}
	    %\mathcal{A}_{p}(\mathbf{u}) = inf \left\{ t | \frac{\#(f \le t \cap \mathcal{N}_{\mathbf{u}})}{\#\mathcal{N}_{\mathbf{u}}} \ge p/100\% \right\},
	    \mathcal{A}_{p}(\mathbf{u}) = inf \left\{ t \mid \frac{\#(\mathcal{A}(\mathcal{N}_{\mathbf{u}}) \le t) }{w^{2}} \ge p/100\% \right\},
	\end{equation}
    where $\mathcal{N}_{\mathbf{u}}$ is a neighborhood of size $w\times w$ centered at $\mathbf{u}$ and $\#(\mathcal{A}(\mathcal{N}_{\mathbf{u}}) \le t)$ is the number of elements of $\mathcal{A}$ evaluated at each point of $\mathcal{N}_{\mathbf{u}}$ that are less than or equal to $t$. Where $\mathcal{N}_{\mathbf{u}}$ extends outside of $\mathcal{A}$, we consider $\mathcal{A}$ to evaluate to $0$. For convenience, in the rest of this paper we use $wf$ to be the $w$/(image width).

	In Section~\ref{sec:results_hyperparams} we demonstrate that both percentile and Gaussian filtering of amplitude spectra lead to state-of-the-art results on different SD tasks, with percentile filtering performing better on average.
	
	%\begin{figure}[ht]
	%\centering
	%\includegraphics[width=0.5\columnwidth]{figs/filter_sal_maps_horiz.pdf}
	%\caption{A demonstration of the parameter effects of Gaussian and percentile filtering on the raw saliency maps (in contrast to the saliency maps in Figure~\ref{fig:gaze_tests}). The original image in (a) contains five figures, four of which are identical. We expect a saliency map to highlight the one unique figure. The Gaussian filter, in (b), appears to do this best with $\sigma > 1$. The percentile filter in (c) appears to do this well for most values of $p$ and $wf > 0.1$, suggesting that a larger window size is important.}
	%\label{fig:filter_sal_maps}
	%\end{figure}

\subsection{Pipeline} 

Our primary approach for solving SD tasks follows a straightforward pipeline.  First, for every training image for a task, we apply a percentile filter to the amplitude spectrum. Second, we train a classifier to predict class labels given the spectra (we discuss the classifiers used in Section~\ref{sec:setup}). In Section~\ref{sec:results_hyperparams} we will examine how the various choices in image transformation affect performance.

\section{Experiments}
\label{sec:results}

\subsection{Setup}
\label{sec:setup}
To evaluate our image preprocessing approach for abstract visual reasoning, we make use of tasks from the SVRT challenge, consisting of 23 binary image classification tasks. As noted by~\citep{stabinger201625}, the SVRT tasks can be split into two groups: spatial relation (SR) problems and same-different (SD) problems. Solving SR problems requires attending to such feature as relative positioning, sizes, alignments, and grouping of figures. The SD problems require comparing individual figures within a single image to identify if they are identical, often under invariants such as scaling or rotation. In particular, we use the task type grouping proposed by~\citep{ricci2018not}, so that there are 9 SD tasks and 14 SR tasks.

We combine our image preprocessing technique with the following few-shot classifiers (with their abbreviated names). \textbf{VGG-pt}: pre-trained VGG16~\citep{simonyan2014very} feature extractor with k-nearest neighbors classifier. \textbf{MAML}: model-agnostic meta-learning~\citep{finn2017model}. \textbf{PN}: Prototypical Networks~\citep{snell2017prototypical}. \textbf{RN}: Relational Networks~\citep{santoro2017simple}. Network architectures, training details, and hyperparameters are described in the supplementary material. \textbf{Fleuret} refers to the Adaboost+spectral features model from~\citep{fleuret2011comparing}, where the only other comparable results on few-shot learning of these tasks is available. The feature type abbreviations are as also follows (with more feature types examined in Section~\ref{sec:results_hyperparams}). \textbf{Raw}: original grayscale images. $\mathcal{A}$: unfiltered amplitude spectra. $\mathcal{A}_{p}$: percentile-filtered amplitude spectra.

\subsection{Results At a Glance}

%A high-level overview of the results can be seen in Table \ref{table:overview}. On the SD tasks, these results clearly demonstrate the superiority of $\mathcal{A}_{p}$ features over not just the raw images, but the unfiltered amplitude spectra, with the performances obtained by the three feature types increasing in increments of approximately 10\% across all classifiers from 50\% to 60\% for $\mathcal{A}$ over raw, and from 60\% to 70\% for $\mathcal{A}_{p}$ over $\mathcal{A}$. On the SR tasks however, we find that $\mathcal{A}$ work best. We look deeper into these results in the rest of this section.

Table~\ref{table:overview} contains a high-level summary of how amplitude spectra filtering makes few-shot learning of SD tasks easier. In particular, we can make the following observations:

\begin{itemize}
    \item By using $\mathcal{A}_{p}$ features for SD tasks, all classifiers tested go from performing at chance-level to above 70\% accuracy with only 10 training samples.
    \item Relational Networks~\citep{santoro2017simple} and Prototypical Networks~\citep{snell2017prototypical} improve the most with $\mathcal{A}_{p}$ features on SD tasks.
    \item When solving SR tasks, $\mathcal{A}_{p}$ features appear to make learning slightly easier, but $\mathcal{A}$ features perform better.
\end{itemize}

% Please add the following required packages to your document preamble:
% \usepackage{booktabs}
% \usepackage{graphicx}
\begin{table}[ht]
\centering
\caption{The test results for each of the four classifiers we use are averages across all SD and SR problems separately and evaluated with the two baseline feature sets (the original image: raw and the unfiltered amplitude spectrum: $\mathcal{A}$).}
\label{table:overview}
\resizebox{0.6\columnwidth}{!}{%
\begin{tabular}{@{}ccrrrl@{}}
\toprule
\textbf{Task type} & \textbf{Feature type} & \multicolumn{1}{l}{\textbf{VGG-pt}} & \multicolumn{1}{l}{\textbf{MAML}} & \multicolumn{1}{l}{\textbf{PN}} & \textbf{RN} \\ \midrule
\textbf{} & raw & 50.7\% & 50.0\% & 50.7\% & 49.6\% \\
\textbf{SD} & $\mathcal{A}$ & 60.5\% & 61.5\% & 63.4\% & 66.6\% \\
\textbf{} & $\mathcal{A}_{p}$ & \textbf{70.1\%} & \textbf{78.4\%} & \textbf{78.7\%} & \textbf{79.1\%} \\ \midrule
\textbf{} & raw & 56.5\% & 51.8\% & 49.3\% & 50.0\% \\
\textbf{SR} & $\mathcal{A}$ & \textbf{58.5\%} & \textbf{59.3\%} & \textbf{62.9\%} & 55.2\% \\
\textbf{} & $\mathcal{A}_{p}$ & 53.4\% & 56.2\% & 55.9\% & \textbf{56.1\%} \\ \bottomrule
\end{tabular}%
}
\end{table}

% While using the raw amplitude does confer a significant benefit by itself (bringing the performance from 50\% to 60\%), filtering doubles the performance increase (from 60\% to 70\%). On the SD tasks however, while the average accuracy using the raw image is higher than for SR, using the unfiltered amplitude performs the best.

% \ref{table:overview}
%TABLE 1:
%- averaged performance of models with baseline feature types for SR and SD

%%%%%%%%%%%%%%%%%%%%%%%%%%%%%%%%%%%%%%%%%%%%%%%%%%%%%%%%%

%\caption{A summary of the test results on the same-different SVRT tasks. Only 10 training samples used for each task. And results are averaged across 10 (most models)) to 20 (our models) trials. On a majority of the tasks where the best performance is measurably above random, our model using self-uniqueness maps outperforms all other models. The only anomaly is problem 16, where the original model by Fleuret et al. exhibits its strongest performance. The large difference between the two version of our model highlights the importance of the self-uniqueness maps on these tasks.}

%\caption{A summary of the test results for all models on the spatial-relation (SR) SVRT tasks. Expectedly, most of the strong baselines are considerable more suited to this set of tasks, improving upon the results in Fleuret et al. by 7\% (the same absolute improvement size of our primary model on the SD tasks). Interestingly, the pretrained VGG16+k-NN model achieves the overall best performance.}

% Please add the following required packages to your document preamble:
% \usepackage{booktabs}
% \usepackage{graphicx}
\begin{table}[ht]
\centering
\caption{A comparison of existing results (Fleuret et al. \cite{fleuret2011comparing}) to four CNN-based few-shot classifiers utilizing $\mathcal{A}_{p}$ features and 10 labelled samples. All four CNN classifiers outperform the existing results on every SD task. On several tasks, our approach is capable of achieving > 95\% accuracy.}
\label{table:sd_results}
\resizebox{\columnwidth}{!}{%
\begin{tabular}{@{}lrrrrrrrrrr@{}}
\toprule
 & \multicolumn{9}{c}{\textbf{SVRT SD Task}} & \multicolumn{1}{l}{} \\ \cmidrule(lr){2-10}
 & \multicolumn{1}{c}{\textbf{1}} & \multicolumn{1}{c}{\textbf{5}} & \multicolumn{1}{c}{\textbf{7}} & \multicolumn{1}{c}{\textbf{15}} & \multicolumn{1}{c}{\textbf{16}} & \multicolumn{1}{c}{\textbf{19}} & \multicolumn{1}{c}{\textbf{20}} & \multicolumn{1}{c}{\textbf{21}} & \multicolumn{1}{c}{\textbf{22}} & \multicolumn{1}{c}{\textbf{Average}} \\ \midrule
Fleuret & 53.0\% & 47.0\% & 47.0\% & 54.0\% & 62.0\% & 51.0\% & 48.0\% & 39.0\% & 53.0\% & 50.4\% \\
VGG-pt & 89.2\% & 70.3\% & 52.2\% & 91.4\% & \textbf{98.8\%} & 55.9\% & 51.9\% & 50.0\% & 70.7\% & 70.1\% \\
MAML & 99.8\% & 92.5\% & 58.4\% & \textbf{100.0\%} & 96.3\% & \textbf{56.9\%} & \textbf{56.0\%} & \textbf{50.6\%} & 95.3\% & 78.4\% \\
PN & 99.6\% & \textbf{95.9\%} & 58.6\% & 99.6\% & 97.3\% & 56.1\% & 55.5\% & 49.6\% & 95.7\% & 78.7\% \\
RN & \textbf{99.9\%} & 94.7\% & \textbf{60.8\%} & 98.7\% & 98.1\% & 56.4\% & 55.3\% & 50.3\% & \textbf{97.6\%} & \textbf{79.1\%} \\ \bottomrule
\end{tabular}%
}
\end{table}
% Please add the following required packages to your document preamble:
% \usepackage{booktabs}
% \usepackage{graphicx}
% \usepackage[table,xcdraw]{xcolor}
% If you use beamer only pass "xcolor=table" option, i.e. \documentclass[xcolor=table]{beamer}
\begin{table}[ht]
\centering
\caption{A comparison on the SR tasks of existing results and the three CNN-based approaches utilizing $\mathcal{A}$ features. MT, PN, and VGG-pt all perform quite close to each other, achieving approximately 8\% better than Fleuret et al. on average.}
\label{table:sr_results}
\resizebox{\columnwidth}{!}{%
\begin{tabular}{@{}lrrrrrrrrrrrrrrr@{}}
\toprule
 & \multicolumn{14}{c}{\textbf{SVRT SR Task}} & \multicolumn{1}{l}{\textbf{}} \\ \cmidrule(lr){2-15}
 & \multicolumn{1}{c}{\textbf{2}} & \multicolumn{1}{c}{\textbf{3}} & \multicolumn{1}{c}{\textbf{4}} & \multicolumn{1}{c}{\textbf{6}} & \multicolumn{1}{c}{\textbf{8}} & \multicolumn{1}{c}{\textbf{9}} & \multicolumn{1}{c}{\textbf{10}} & \multicolumn{1}{c}{\textbf{11}} & \multicolumn{1}{c}{\textbf{12}} & \multicolumn{1}{c}{\textbf{13}} & \multicolumn{1}{c}{\textbf{14}} & \multicolumn{1}{c}{\textbf{17}} & \multicolumn{1}{c}{\textbf{18}} & \multicolumn{1}{c}{\textbf{23}} & \multicolumn{1}{c}{\textbf{Average}} \\ \midrule
Fleuret & 55.0\% & 54.0\% & 56.0\% & 50.0\% & 57.0\% & 52.0\% & 50.0\% & 52.0\% & 46.0\% & 50.0\% & 51.0\% & 59.0\% & 50.0\% & 53.0\% & 52.5\% \\
VGG-pt & 66.2\% & 50.5\% & \textbf{66.6\%} & \textbf{51.3\%} & 75.0\% & 50.7\% & 71.2\% & 50.7\% & \textbf{57.2\%} & 53.7\% & \textbf{67.6\%} & 53.2\% & 50.8\% & \textbf{54.8\%} & 58.5\% \\
MAML & 68.9\% & 56.1\% & 61.4\% & 50.3\% & 73.8\% & 51.2\% & 74.6\% & 60.1\% & 54.2\% & 65.1\% & 58.9\% & 51.9\% & 51.3\% & 52.0\% & 59.3\% \\
PN & \textbf{77.8\%} & \textbf{58.2\%} & \textbf{66.6\%} & 50.8\% & \textbf{83.4\%} & \textbf{51.4\%} & \textbf{83.7\%} & \textbf{64.1\%} & 54.8\% & \textbf{70.4\%} & 59.8\% & \textbf{53.3\%} & \textbf{53.9\%} & 52.0\% & \textbf{62.9\%} \\
RN & 67.8\% & 51.2\% & 53.9\% & 51.1\% & 57.1\% & 51.1\% & 67.1\% & 52.8\% & 51.6\% & 62.1\% & 53.4\% & 51.6\% & 51.0\% & 50.4\% & 55.2\% \\ \bottomrule
\end{tabular}%
}
\end{table}

\subsection{A Closer Look}
\label{sec:obs_sd}

% \ref{table:sd_results}
%TABLE 2:
%- performance of models on SD tasks with the filtered features

%A comparison of existing results (Fleuret et al. \citep{fleuret2011comparing}) to three CNN-based classifiers utilizing $\mathcal{A}_{p}$ features and 10 labelled samples. All three CNN classifiers outperform the existing results on every SD task, with the Prototypical Network approach achieving the highest average accuracy. Interestingly, despite the VGG16 network being pretrained on natural images, the features extracted from the filtered amplitude spectra allow the model to perform better than Mean Teacher, and on problem 16, allows it to achieve significantly better accuracy than the more advanced approaches. On several tasks (1, 15, 16, 22) our models are capable of achieving test accuracies > 90\%

\textbf{Same-different tasks.} Perhaps the most immediate observation from Table~\ref{table:sd_results} is the contrast in performance between classifiers utilizing $\mathcal{A}_{p}$ features and the existing results. On every task, our approaches outperform Fleuret et al., with the best classifier, Relational Networks, achieving an accuracy nearly 30\% higher on average. 
Two problems where our models perform particularly well are \#1 and \#15. In class 1 for both tasks, some number of identical shapes are present (2 and 4 respectively), while in class 2, the shapes are all unique. These represent the most purely same-different problems. The highest performance of Fleuret et al. is achieved on \#16  -- in this task, both classes contain six identical shapes, but in class 1, the shapes on one side of the image can be obtained from those on the other side by reflection about the vertical image bisector. In class 2, the positions of the shapes are reflected, but not the details of the shapes themselves\footnote{Descriptions of each SVRT problem can be found in the appendix for~\citep{fleuret2011comparing}, found at \url{https://www.pnas.org/content/pnas/suppl/2011/10/12/1109168108.DCSupplemental/Appendix.pdf}}.

% \ref{table:sr_results}
%TABLE 3: 
%- performance of models on SR tasks with default fft features
    
\textbf{Spatial relation tasks.} Table~\ref{table:sr_results} contains the results using the best-performing feature type on these tasks, $\mathcal{A}$, according to Table~\ref{table:overview}. Aside from when using the RN classifier, this feature type performs similarly for both SD and SR problems, and consistently better than the raw images. This suggests that while $\mathcal{A}$ is not very useful for CNNs in providing uniqueness-type information, these features contain information difficult for CNNs to extract from raw images. On these tasks, the spectral features allow us to achieve up to 10\% higher on average than Fleuret et al. with the PN classifier. Previous work has shown that using the raw images, CNNs can achieve very high accuracies on the SR tasks when given a large amount of training data~\citep{stabinger201625, ricci2018not}, but in this few-shot case we observe that the raw images only achieve up to 56.5\% accuracy with the VGG-pt classifier, lower than the high-90s averages reported by~\citep{stabinger201625} with 20,000 training samples and~\citep{ricci2018not} with several million training samples. This demonstrates that improving few-shot performance on the SVRT SR tasks is worth further study.

%%%%%%%%%%%%%%%%%%%%%%%%%%%%%%%%%%%%%%%%%%%%%%%%%%%%%%%%%5
\subsection{Effects of Hyperparameters}
\label{sec:results_hyperparams}

% Please add the following required packages to your document preamble:
% \usepackage{booktabs}
% \usepackage{graphicx}
\begin{table}[ht]
\caption{A comparison of the SD task performance of various feature types considered. $\mathcal{A}_{g}$ refers to the Gaussian-filtered amplitude spectra and $\mathcal{S}_{p}$ refers to the saliency map produced by taking the inverse Fourier transform with the original phase spectrum and the percentile-filtered amplitude.}
\label{table:feature_compare}
\resizebox{\columnwidth}{!}{%
\begin{tabular}{@{}lrrrrrrrrrr@{}}
\toprule
 & \multicolumn{9}{c}{\textbf{SVRT SD Task}} & \multicolumn{1}{l}{\textbf{}} \\ \cmidrule(lr){2-10}
 & \multicolumn{1}{c}{\textbf{1}} & \multicolumn{1}{c}{\textbf{5}} & \multicolumn{1}{c}{\textbf{7}} & \multicolumn{1}{c}{\textbf{15}} & \multicolumn{1}{c}{\textbf{16}} & \multicolumn{1}{c}{\textbf{19}} & \multicolumn{1}{c}{\textbf{20}} & \multicolumn{1}{c}{\textbf{21}} & \multicolumn{1}{c}{\textbf{22}} & \multicolumn{1}{c}{\textbf{Average}} \\ \midrule
raw & 51.0\% & 49.5\% & 50.3\% & 52.5\% & 50.8\% & 50.3\% & 49.8\% & 50.8\% & 50.9\% & 50.7\% \\
$\mathcal{A}$ & 63.3\% & 66.1\% & 54.2\% & 76.5\% & 58.9\% & 53.1\% & 51.6\% & 49.0\% & 71.7\% & 60.5\% \\
$\mathcal{A}_{p}$ & \textbf{89.2\%} & \textbf{70.3\%} & 52.2\% & \textbf{91.4\%} & \textbf{98.8\%} & 55.9\% & 51.9\% & 50.0\% & 70.7\% & \textbf{70.1\%} \\
$\mathcal{A}_{g}$ & 64.6\% & 63.7\% & \textbf{58.3\%} & 85.8\% & 67.2\% & \textbf{57.1\%} & \textbf{53.5\%} & \textbf{50.8\%} & \textbf{77.5\%} & 64.3\% \\
$\mathcal{S}_{p}$ & 86.5\% & 62.4\% & 52.8\% & 88.4\% & 52.1\% & 52.1\% & 49.8\% & 50.4\% & 69.0\% & 62.6\% \\ \bottomrule
\end{tabular}%
}
\end{table}

%\textbf{$\mathcal{A}$}}
%\textbf{$\mathcal{A}_{p}$
%\textbf{$\mathcal{A}_{g}$}
%\textbf{$\mathcal{S}_{p}$}

\textbf{Choice of image transformation.} In Table~\ref{table:feature_compare} we provide support for our choice of percentile-filtered amplitude spectra as the primary image preprocessing approach in our experiments. For these comparisons, we report the results of the VGG-pt classifier (chosen for its speed, as no neural networks require training). While both filtering methods (percentile and Gaussian) achieve the best performance on subsets of the SD tasks, percentile filtering achieves a 6\% higher average accuracy than Gaussian filtering. Additionally, while it may be more intuitive to reconstruct the images with the filtered spectra before training the classifiers (feature type $\mathcal{S}_{p}$), Table~\ref{table:feature_compare} demonstrates that this achieves 8\% lower than the $\mathcal{A}_{p}$, but still better than the raw images and the unfiltered amplitude spectra. However, we note that constructing saliency maps is often subject to additional post-processing parameters which we did not tune.

%Our primary image preprocessing technique uses the percentile-filtered amplitude spectra instead of Gaussian filtering which was used in \citep{li2013visual} for producing saliency maps. Additionally, instead of reconstructing the images with the filtered spectra, we provide $\mathcal{A}_{p}$ directly to the CNN classifiers. Both of these choices are not immediately obvious. In Table \ref{table:feature_compare}, we demonstrate the performance of the alternatives when used with the transfer-learning classifier. This classifier was chosen for these experiments for its speed (Mean Teacher and Prototypical Networks both require training). We observe that the worst performing feature type is still the raw images. The Gaussian filtered amplitude spectra ($\mathcal{A}_{g}$) performs second-best to percentile filtering, but still significantly lower. Finally, the saliency map ($\mathcal{S}_{p}$) calculated using $\mathcal{A}_{p}$ performs even worse, but still over 10\% better than the raw images. However, we note that constructing saliency maps is often subject to additional post-processing parameters which we did not tune. %Interestingly, $\mathcal{A}_{g}$ outperforms $\mathcal{A}_{p}$ on almost exclusively the most difficult tasks (\#7, \#19, \#20, and \#21).

\textbf{Filter parameters.} We found that the most important parameter for the percentile filter was the size of the neighborhood around each point, $wf$, used to calculate the p-percentile, with larger values performing better. Unfortunately, calculating the p-percentile on a window requires first ranking every value, making the calculation slow for large windows. Thus, while we only tried $wf$ values up to 0.2, we strongly suspect that using larger values would improve performance on the SD tasks by several percent. For the $p$ value, we found 10 to generally perform best during hyperparameter tuning on the validation set, but with only minor performance decreases when 5 or 20 were used. When tuning the Gaussian kernel to produce the results in Table~\ref{table:feature_compare}, we found the optimal value for $\sigma$ to sharply increase at 2, and slowly fall off for larger values.

\section{Conclusion}
\label{sec:conclusion}

We have presented an image preprocessing technique allowing few-shot deep learning classifiers to achieve improved accuracy on same-different (SD) problems. SD problems are a fundamental type of visual reasoning task often trivial for humans to solve with few samples while deep learning approaches training on millions of samples have been unsuccessful. Discovering machine learning approaches capable of solving these tasks is valuable in working towards automating highly abstract human skills.

To solve SD problems, we propose training CNN-based classifiers on the percentile-filtered amplitude spectra. As has been previously established with Gaussian filters, filtering these spectra correspond to removing the non-unique parts of an image. In line with previous work suggesting that fully convolutional approaches have difficulty learning to solve SD tasks, we demonstrate that a variety of state-of-the-art few-shot classifiers achieve only 50\% binary classification accuracy on SD problems when trained on raw images. However, combining the classifiers with our image preprocessing technique allows them to achieve between 70\% and 80\% accuracy, outperforming the existing comparable state-of-the-art on SVRT SD tasks in the few-shot case, solidifying the effectiveness of this approach.
%Additionally, we show that state-of-the-art CNN-based few-shot classifiers exhibit poor performance on this task when training with raw images, which concurs with previous work suggesting that fully convolutional approaches have difficulty learning to solve same-different tasks.

%Close observation of the performance of our approach on the SVRT tasks suggests an avenue for improvement in future work. We identify a subset of the same-different tasks that are especially difficult and suggest that this comes from requiring similarity comparisons under rotational, reflection, and scaling invariance. Finding an efficient way to incorporate such invariants into saliency maps would likely lead to clear performance on these very hard problems. Progress on these tasks may also lead to more robust performance at real-world instantiations of same-different tasks.

% This demonstrates that improving few-shot performance on the SVRT SR tasks is also worth further study.

\bibliographystyle{unsrt}
\bibliography{neurips_2019}

\newpage
\appendix
\appendixpage

\section{Network architecture and hyperparameters}
\label{sec:supplemental}
% see supp for https://papers.nips.cc/paper/7404-image-to-image-translation-for-cross-domain-disentanglement

Here we discuss the implementation details and hyperparameters used for each model in our experiments. For all models, we use SVRT images of size 96x96, and for each SVRT task, we train models with 10 labelled samples (to compare to the few-shot experiments done in \citep{fleuret2011comparing}) and evaluate on 1000 samples, where performance is measured with classification accuracy. To produce all experimental results, we average over 10 trials. For hyperparameter tuning, we average across 5 trials for each measurement on only the SD tasks, using 1000 validation images for each task. To produce the test results, we average across 10 trials, with 1000 different unseen test images each trial. For model training, we use 10 new images each trial. The code to generate samples for SVRT tasks is publicly available at \url{http://www.idiap.ch/~fleuret/svrt/}. Next, we discuss the details of tuning each classifier.

    \textbf{Transfer Learning.} For this model, we used the VGG16 architecture pre-trained on ImageNet \citep{simonyan2014very} and provided through Keras \citep{chollet2015}. We extract features from the last set of convolutions. For the classifier, we use k-nearest neighbors. The value for $k$ was chosen from $\{1, 3, 5\}$ and the best value was found to be $k=1$ for all feature types.

    % 4, 8, 16, 32, 64
    \textbf{Prototypical Networks.} We use a publicly available implementation (\url{https://github.com/orobix/Prototypical-Networks-for-Few-shot-Learning-PyTorch}) with the same architecture for the embedding stage as in \citep{snell2017prototypical} and \citep{lake2011one}. Three training samples are randomly chosen from each class for its support set. The optimal number of epochs was chosen from $\{8, 16, 32, 64\}$ to maximize the average score on validation SD problems. When training on raw images, 32 epochs was found to perform best, and 64 epochs was found to perform best when using the spectral features.
    % link:https://github.com/orobix/Prototypical-Networks-for-Few-shot-Learning-PyTorch/tree/master/src

    \textbf{MAML.} We use a publicly available implementation (\url{https://github.com/katerakelly/pytorch-maml}) with the same architecture and hyperparameters as in the supervised Omniglot experiments in \citep{finn2017model}. However, we choose the number of training epochs from ${8, 16, 32, 64}$ to maximize the average score across validation SD problems. When training on raw images, 8 epochs was found to perform best, and 32 epochs was found to perform best when using the spectral features.
 
    \textbf{Relational Networks.} We use a publicly available implementation (\url{https://github.com/kimhc6028/relational-networks}) with the same architecture used for CLVER in \citep{santoro2017simple}. However, since this model is originally designed for tasks which contain images paired with textual questions, we modify the model by removing components associated with question processing so it can be applied to SVRT tasks. We select the number of training epochs from $\{64, 128, 256\}$ and learning rate from $\{1\text{e-}4, 2.5\text{e-}4, 1\text{e-}5\}$. When training on raw images, 128 epochs and learning rate of $1\text{e-}4$ was selected. When training on spectral features, 256 epochs and a learning rate of $5\text{e-}4$ were selected.

    When optimizing the parameters for the percentile-filter for each classifier, we chose $p$ from $\{5, 10, 20, 40\}$ and $wf$ from $\{0.05, 0.1, 0.2\}$. The optimal values for $p$ and $wf$ were identified to be 10 and 0.2 respectively for all classifiers, except for the transfer learning model, which used $p = 10$ and $wf = 0.2$.

\end{document}